\documentclass[12pt,a4paper,twocolumn]{article}

\usepackage[utf8]{inputenc}
\usepackage[english]{babel}
\usepackage{alphabeta}
\usepackage{amsmath}
\usepackage{fancyhdr}
\usepackage{titlesec}
\usepackage{abstract}
\usepackage{setspace}
\usepackage{url}
\usepackage{authblk}
\usepackage[inline]{enumitem}

\usepackage[square,sort,comma,numbers]{natbib}
\titleformat{\section}
  {\normalfont\large\bfseries}
  {\thesection.}{.5em}{}
  
\renewcommand{\vec}[1]{\mbox{\boldmath$#1$}}

\newcommand{\myemph}[1]{{\emph{#1}}}
\newcommand{\myeg}{{\emph{e.g., }}}
\newcommand{\myie}{{\emph{i.e., }}}
\newcommand{\myetal}{{\emph{et al. }}}

\DeclareMathOperator{\argmax}{arg\,max}
\DeclareMathOperator{\argmin}{arg\,min}

\makeatletter
\renewenvironment{abstract}{
   \begin{center}
     {\bfseries \normalsize\abstractname\vspace{\z@}}
   \end{center}
   \footnotesize}
{\endquotation}
\makeatother

\date{}
\begin{document}
\title{Ontology-based Feature Selection: A Survey}
\author{Konstantinos Sikelis \\ cti20004@ct.aegean.gr \and George E.Tsekouras \\ gtsek@ct.aegean.gr \and Konstantinos Kotis \\ kotis@aegean.gr}
\affil{\small Intelligent Systems Lab, Dept. of Cultural Technology and Communication, University of the Aegean, Mytilene, 81100, Greece}

\author{User1 \\ user1@mailserver.xx \and User2 \\ user2@mailserver.xx}

\twocolumn[
\maketitle

\begin{@twocolumnfalse}
\begin{abstract}
\begin{singlespace}
The Semantic Web emerged as an extension to the traditional Web, towards adding meaning to a distributed Web of structured and linked data. 
At its core, the concept of ontology provides the means to semantically describe and structure information and data and expose it to software and
human agents in a machine and human-readable form. For software agents to be realized, it is crucial to develop powerful artificial intelligence and machine 
learning techniques, able to extract knowledge from information and data sources and represent it in the underlying ontology. This survey aims to 
provide insight into key aspects of ontology-based knowledge extraction, from various sources such as text, images, databases and human expertise,
with emphasis on the task of feature selection. First, some of the most common classification and feature selection algorithms
are briefly presented. Then, selected methodologies, which utilize ontologies to represent features and perform feature selection and classification, 
are described. The presented examples span diverse application domains, \myeg medicine, tourism, mechanical and civil engineering, and demonstrate the 
feasibility and applicability of such methods.
\end{singlespace}
\end{abstract}
\vspace{1.0cm}
\end{@twocolumnfalse}]

\section{Introduction}
The recent development of the Semantic Web enables the systematic representation of vast amounts of knowledge within an ontological
framework. An ontology is a formal and explicit description of shared and agreed knowledge as a set of concepts within a domain and the relationships that hold among them. The ontological model provides a rich set of axioms to link pieces of information and enables automatic reasoning to infer knowledge that has not been explicitly asserted before.
 
In many cases, reasoning can be cast as a classification task. An important step towards an accurate and efficient classification, is feature selection. Consequently, identification of high-quality features from an ontology hierarchy plays a significant role in the ability to extract information from 
an ontological model. 

This report summarizes related work, which tackles the problem of feature representation and selection based on ontologies in the context of knowledge 
extraction from documents, images, databases and human expertise. In the first and second sections, a brief summary of selected classification and feature selection methods is presented. In the third section, the concept of ontology as a building block of the Semantic Web is introduced. In the fourth section, example methodologies of ontology-based feature selection are summarized. In the fifth section open issues and challenges are considered. In the last section the conclusions of this survey are discussed.

\section{Data Classification}
One of the most common applications of machine learning is data classification. Data classification can be defined as the data analysis task which, 
given a set of observations belonging to known categories, aims at identifying to which category a new observation belongs. In the case that the 
feature and target variables are not categorical, but take continuous values, the classification task is called regression. 

In essence this problem attempts to learn the relationship between a set of feature variables and a target variable of interest. In practice a 
large variety of problems can be expressed as associations between feature and target variables which provides a broad range of applications such as 
customer target marketing \cite{REF1, REF2}, medical disease diagnosis \cite{REF3, REF4, REF5}, speech and handwriting recognition \cite{REF6, REF7, REF8, REF9}, multimedia data analysis \cite{REF10, REF11}, biological data analysis \cite{REF12}, 
document categorization and filtering \cite{REF13, REF14}, and social network analysis \cite{REF15, REF16, REF17}. 

Classification algorithms typically contain two steps, the learning step and the testing step. The first one constructs the classification model, while 
the second evaluates it by assigning class labels to unlabeled data. For a test instance under consideration, the output of a classification algorithm 
may be presented either as discrete label or numerical score. In the former case, 
the classifier assigns a single label, which identifies the class of the test instance, while in the latter a numerical score is returned which associates 
the test instance with each class. The advantage of a numerical score is that it incorporates the additional information of ``belongingness'' of the test 
instances to each category and thus it facilitates their ranking.

A close relative to the classification problem is data clustering \cite{REF18, REF19}. Clustering is the task of dividing a population of data points into a number of 
groups, such that the members of the same group are in some sense similar to each other and dissimilar to the data points in other groups. The key 
difference between the two tasks is that in the case of clustering, data are segmented using similarities between feature variables, while in the 
case of classification, data partitioning is based on a training data set. Consequently, clustering has no understanding of the underlying group 
structure, whereas classification uses knowledge encoded in the training data set in the form of a target variable. As a result, the classification 
task is referred to as supervised learning and clustering as unsupervised learning. 

A plethora of methods can be used for data classification. Some of the most common are probabilistic methods \cite{REF20, REF21, REF22}, decision trees \cite{REF23, REF24, REF25}, rule-based methods \cite{REF26, REF27, REF28},
Support Vector Machine methods \cite{REF29, REF30}, instance-based methods and neural networks \cite{REF31, REF32}.

Probabilistic methods are based on two probabilities, namely a prior probability, which is derived from the training data, and a posterior probability
that a test instance belongs to a particular class. There are two approaches for the estimation of the posterior probability. In the first approach, 
called generative, the training dataset is used to determine the class probabilities and class-conditional probabilities and the Bayes theorem is employed 
to calculate the posterior probability. In the second approach, called discriminative, the training dataset is used to identify a direct mapping of a test 
instance onto a class.

A common example of a generative model is the \myemph{Naive Bayes classifier} \cite{REF31, REF32}. Assuming a test instance T with features 
\mbox{$\vec{x}=[x_1, x_2, \dotsc, x_d]$}, the probability that T belongs to class y can be calculated with the Bayes theorem:
\begin{equation}P(y|\vec{x})=P(y) P(\vec{x} | y)/P(\vec{x})\end{equation}
Then the problem of classification is to find the class which maximizes the above probability given the features of the test instance T. Since the 
denominator is constant across all classes the problem can be approximated as follows:
\begin{equation}P(y|\vec{x}) \propto P(y) P(\vec{x} | y) \end{equation}
In the above equation, the class probability \mbox{$P(y)$} is the fraction of training instances which belong to class y. The class-conditional probability 
\mbox{$P(\vec{x}|y)$} can be calculated under the naive Bayes assumption that the features \mbox{$x_i$} are independent to each other. This simplification 
allows the class-conditional probability to be calculated as a product of the feature-wise conditional probabilities:
\begin{equation}P(\vec{x}|y) = \prod_{j=1}^{d} P(x_j | y) \end{equation}
The term \mbox{$P(x_i|y)$} is computed as the fraction of the training instances classified as y, which contain the ith feature. Generally, the naive Bayes 
assumption does not hold, however the naive Bayes model has been proven to work quite well in practice. 

A popular discriminative classifier is the \myemph{logistic regression} \cite{REF31}, where the posterior probability is modeled as:
\begin{equation}P(y|\vec{x})=(1+e^{-\vec{\theta}^T\vec{x}})^{-1} \end{equation}
where \mbox{$\vec{\theta}$} is a vector of parameters to be estimated from the training data. Given m independent training instances with class labels 
\mbox{$\vec{y}=[y_1,y_2,\dotsc,y_m]$} and feature vectors \mbox{$X=[\vec{x_1}, \vec{x_2},\dotsc,\vec{x_m}]$} respectively, the unknown parameters are 
derived from the maximization of the posterior probability with respect to \mbox{$\vec{\theta}$}: 
\begin{equation}\vec{\theta}\leftarrow \argmax_\theta P(\vec{y}|X)\end{equation}

In \myemph{Decision Tree Classification} \cite{REF20, REF21, REF22}, data are recursively split into smaller subsets until all formed subsets exhibit class purity \myie 
all members of each subset are sufficiently homogeneous and belong to the same unique class. In order to optimize the decision tree, an impurity measure 
is employed and the optimal splitting rule at each node is determined by maximizing the amount that the impurity decreases due to the split. 
A commonly used function for this purpose is the Shannon entropy. If T are the training records at the Nth node of the decision tree and 
\mbox{$(p_1, p_2, \dotsc, p_k)$} are the fractions of the records which belong to k different classes, then the entropy H(T) is: 
\begin{equation} H(T) = -\sum_{i=1}^k {p_i\,log(p_i)} \end{equation}
and the Information Gain (impurity decrease), from picking attribute \mbox{$\alpha$} for splitting the node, is: 
\small\begin{equation}\begin{split}IG(T,\alpha)&=H(T)-H(T|\alpha)\\&= H(T)-\sum_{v\in{vals(\alpha)}}{\frac{|T_v|}{|T|}H(T_v)}\end{split}\end{equation}\normalsize
where \mbox{$T_v$} denotes the set of all training records where attribute \mbox{$\alpha=v$}.

An extension to decision tree classification is the \myemph{Random Forest (RF)} algorithm \cite{REF33}. This algorithm, essentially, trains a large set of decision trees 
and combines their predictive ability in a single classifier. These decision trees vote for the class membership of an instance and the class with 
the majority of votes becomes the result of the classification. In order to build each decision tree, first a training data set is derived from the original 
dataset, by random sampling with replacement, meaning that training records are allowed to be selected multiple times. Then, during training, only a random 
subset of features, out from the whole feature space, is considered for splitting each node. These two steps introduce randomness into the tree
learning process and increase the diversity of the base classifiers, thus improving the overall accuracy of the classification process. The RF
classifier belongs to a broader family of methods called \myemph{ensemble learning} \cite{REF31}.

A classification method closely related to Decision Trees is called \myemph{Rule Based Classification} \cite{REF26, REF27, REF28}. Essentially, all paths in a decision tree represent 
rules, which map test instances to different classes. However, for Rule-based methods the classification rules are not required to be disjoint, rather they 
are allowed to overlap. Rules can be extracted either directly from data (rule induction) or built indirectly from other classification models.  

CN2 \cite{REF34} and RIPPER \cite{REF35} are two popular direct algorithms, which use the sequential covering paradigm. In this approach, a target class is selected and the 
corresponding rule is mined from the data incrementally by successive additions of conjuncts to the rule antecedent, 
according to a certain criterion (\myeg maximization of FOIL’s Information gain for RIPPER or minimization of the entropy measure for CN2). After each 
rule is grown, all matched training instances are removed. This process is repeated until all remaining training tests belong to one class. This class 
serves as the default, and is assigned to all test instances which do not activate any rules.

A novel family of algorithms which aim at mining classification rules indirectly, is the so called \myemph{Associative Classification} \cite{REF36}. Associations are 
interesting relations between variables in large datasets. Association rules can quantify such relations by means of constraints on measures of 
significance or interest. The best-known constraints are imposing minimum threshold values on support and confidence. In the training phase, an 
associative classifier, mines a set of Class Association Rules  from the training data. This process occurs in two steps. First, the training data 
set is searched for repeated patterns of feature-value pairs. Then, association rules are generated from the resulting pairs, according to the 
proportion of the data they represent (support) and their accuracy (confidence). The mined CARs are used to build the classification model according 
to some strategy such as applying the strongest rule, selecting a subset of rules, forming a combination of rules or using rules as features. 

Another promising new class of methods are \myemph{Support Vector Machines (SVM)} \cite{REF37}. SVM classifiers are generally defined for binary classification tasks. 
Intuitively, they attempt to draw a decision boundary between the data items of two classes, according to some optimality criterion. A common criterion 
employed by SVM is that the decision surface must be maximally far away from any data point. The margin of the separation is determined by the distance 
from the decision surface to the closest data points. Such data points are called support vectors. 

Finding the maximum margin hyperplane is a quadratic optimization problem \cite{REF38}. In case the training data are not linearly 
separable, slack variables can be introduced in the formulation to allow some training instances to violate the support vector constraint, \myie 
they are allowed to be on the “other” side of the support vector from the one which corresponds to their class. When data are non-linear, a linear model 
will not have acceptable accuracy. However, in this case it is possible to use a mapping to transform the original data into linear-separable data in a 
higher dimension. Such transformations rely on the notion of similarity between data records. Such similarities are expressed as dot-products which are 
called kernel functions. A typical function of this kind is the Gaussian Kernel function.

Recently, \myemph{Artificial Neural Networks} have been proven to be powerful classifiers \cite{REF32}. They attempt to mimic the human brain by means of an 
interconnected network of simple computational units, called neurons. Neurons are functions which map an input feature vector to an output value according 
to predefined weights. These weights express the influence of each feature over the output of the neuron and are learned during the training phase.
A typical tool to perform the training process is the back-propagation algorithm. Backpropagation uses the chain rule to compute the derivative of the 
error (loss function) with respect of the weights of the network and applies gradient methods (\myeg stochastic gradient descent) to find weight 
values which minimize the error. 

In general, the neural mapping, results from the composition of a net value function \mbox{$\xi$}, which summarizes the input data into a net value 
\mbox{$v=\xi(\vec{x},\vec{w})$} and an activation function which transforms the net value into an output value \mbox{$h=\phi(v)$}. Depending on the form of 
functions \mbox{$\xi$} and \mbox{$\phi$}, several neuron types can be derived. For example, the perceptron or linear neuron uses a weighted sum of the 
features plus a bias value as the net value function while the activation function is a simple multiple of the net value with a parameter \mbox{$\alpha$}, 
usually equal to 1. The sigmoidal neuron uses the same net value function as the perceptron, but replaces the activation function either with the 
sigmoid function or with the hyperbolic tangent function. On the other hand, the distance neuron uses the same activation function as the perceptron, 
but replaces the net value function with a distance measure \mbox{$\xi(\vec{x},\vec{w})=\parallel{\vec{x}-\vec{w}}\parallel$}. The radial basis function
neuron uses again a distance measure, typically the Euclidean distance or the Mahalanobis distance, as the net value function but employs the exponential 
\mbox{$\phi=exp(-v)$} as the activation function. Polynomial units extend the perceptron net value function by adding a quadratic form, 
\mbox{$\xi(\vec{x},\vec{w})=\vec{w}^T\vec{x} + w_0+\vec{x}^T W \vec{x}$}, while the activation function is usually linear or sigmoid.

The aforementioned classifiers are eager learners, meaning that they build the generalization function during training by observing the whole training set, 
but without any insight into future queries. When the training data are not distributed evenly in the input space, eager methods  may suffer from poor 
generalization capabilities. Essentially, by trying to minimize the global error over the whole training set they may exhibit inaccurate approximations 
of local subspaces. 

Instance-based classification attempts to parry this shortcoming \cite{REF39}. In the first phase, Instance-based classifiers, do not build any approximation models, rather they simply store the training records. When a query 
is submitted, the system uses a distance function to extract, from the training data set, those records which are most similar to the test instance. Label 
assignment is performed based on the extracted subset. Common, instance-based classifiers are the K-Nearest neighbor (KNN), Self-organizing Map (SOM), Learning Vector Quantization (LVQ) \cite{REF40}. A generalization of instance-based learning is lazy learning, where training examples in the neighborhood of the test instance are used to train a locally optimal classifier. 
The field of classification is vast and still in its infancy. For an excellent in depth discussion on classification methods, the curious reader is referred to \cite{REF31}. 

\section{Feature Selection}
The first step towards successful classification, is finding an accurate representation of the problem domain. In other words, one has to concretely 
define the features and target classes, which will be input to the classifier. In practice, these are part of available raw data, for example in huge 
databases or through a live data collection system. In this pile of data, explicit features and target classes are not so clear cut, and need to be 
uncovered. In general, this process is called Feature Engineering (FE) and encompasses algorithms for generating features from raw data (feature generation), 
transforming existing features (feature transformation), selecting most important features (feature selection), understanding feature behavior 
(feature analysis) and determining feature importance (feature evaluation) \cite{REF41}.

Feature selection is one of the most popular and well studied methods of FE. When data records are derived from databases, usually they contain data 
columns which are unrelated to the targets under consideration. If all data are blindly used as features they increase the dimensionality of the 
problem, which has negative impact on the classification task. First of all, an increasing number of dimensions in the feature space, results in 
exponential expansion of the computational cost. This issue is directly related to the curse of dimensionality problem. Furthermore as the volume of 
feature space increases, it becomes sparsely populated and even close data points may be driven apart from irrelevant data, thus appearing as far away as 
unrelated data points. This will increase overfitting and reduce the accuracy of the classifier. Also, the more features are used for the classification, 
the more difficult it becomes to decode the underlying relationship of the used features to the classifier. Restricting the used features to only those 
that are strictly relevant to the target classes results in improved interpretability of the model. Finally, data collection becomes easier and with higher 
quality when the number of features is small.  

The feature selection process attempts to remedy these issues by identifying features which can be excluded without adversely affecting the classification 
outcome. Feature selection is closely related to feature extraction. The main difference is that while feature selection maintains the physical meaning 
of the retained features, feature extraction attempts to reduce the number of dimensions by mapping the physical feature space on a new mathematical space. 
Consequently, the derived features lack a physical meaning, which makes the addition and analysis of new physical features problematic. In this sense 
feature selection appears to be superior in terms of readability and interpretability.

Feature selection can be supervised, unsupervised or semi-supervised. Supervised methods consider the classification information, and use measures to 
quantify the contribution of each feature to total information, thus keeping only the most important ones. Unsupervised methods attempt to remove 
redundant features in two steps. First, features are clustered into groups, using some measure of similarity, and then the features with the strongest 
correlations to the other features in the same group are retained as the representatives of the group. Identification and removal of irrelevant features 
is more difficult and abstract and depends on some heuristic of relevance or interestingness. To devise such heuristics researchers have employed several 
measures such as scatter separability, entropy, category utility, maximum likelihood, density, and consensus \cite{REF42}. Semi-supervised feature selection 
addresses the case when both a large set of unlabeled and a small set of labeled data are available. The idea is to use the supervised class-based clustering 
of features in the small dataset as constraint for the unsupervised locality-based clustering of the features in the large dataset. 

Depending on whether and how they use the classification system, feature selection algorithms, are divided into three categories, namely filters, 
wrappers and embedded models.  

Filter models select subsets of variables as a preprocessing step, independently of the chosen classifier. In the first step, features are analyzed 
and ranked on the basis of how they correlate to the target classes. This analysis can either consider features separately and perform ranking 
independently of the feature space (univariate), or evaluate groups of features (multivariate). Multivariate analysis has the advantage that 
interactions between features are taken into account in the selection process. In the second step, the features with the highest scores are used in 
the classification model. 

Some of the most common evaluation metrics which have been used for ranking and filtering are:

\myemph{Chi-Square}: The \mbox{$\chi^2$} correlation uses the contingency table of a feature target-pair to evaluate the likelihood that a selected feature 
and a target class are correlated. The contingency tables shows the distribution of one variable (the feature) in rows and another (the target) in columns. 
Based on the entries the observed values are calculated. Also under the assumption that the variables are independent (null hypothesis) the expected 
values are derived. Small values of \mbox{$\chi^2$} show that the expected values are close to the observed values, thus the null hypothesis stands. On the 
contrary high values show strong correlation between the feature and the target value. The \mbox{$\chi^2$} metric is defined as
\small\begin{equation}\chi^2=\sum \frac{(O_{ij}-E_{ij})^2}{E_{ij}}\end{equation}\normalsize
where \mbox{$O_{ij}$} and \mbox{$E_{ij}$} the observed and expected values of feature $i$ with respect to class $j$. 

\myemph{ANOVA}: A metric related to \mbox{$\chi^2$} is Analysis of Variance. It tests whether several groups are similar or different by comparing their 
means and variances, and returns an F-statistic which can be used for feature selection. The idea is that a feature where each of its possible values 
corresponds to a different target class, will be a useful predictor. Let \mbox{$\bar{f}$} be the mean value of feature $f$ (grand mean),  
\mbox{$\bar{f_i}$} the mean value of feature $f$ in each individual group $i$ with class assignement \mbox{$c_i$} and \mbox{$f_{ij}$} the value of feature $f$ 
at record $j$ of \mbox{group $i$}. The Sum of Squares Between is defined as \mbox{$SSB=\sum{(\bar{f}-\bar{f_i})^2}$} and the Sum of Squares Within is 
\mbox{$SSW=\sum{(f_{ij}-\bar{f_i})^2}$}.Then the F-statistic is computed as:
\small\begin{equation} F = \frac{SSB/DF_B}{SSW/DF_W} \end{equation}\normalsize
where \mbox{$DF_B=\mbox{NumberOfClasses}-1$} and \mbox{$DF_W=\mbox{NumberOfGroupRecords}-1$} are the degrees of freedom of the target group and feature group 
respectively. A high value of $F$ indicates that feature $f$ has high predictive power. 

\myemph{Fisher Score}: It is based on the intuition that high quality features should assign similar values to instances in the same class and different 
values to instances from different classes. Let \mbox{$\mu_i$} denote the mean of the $i$-th feature, \mbox{$n_j$} the number of instances in the $j$-th class 
and \mbox{$\mu_{ij}, \rho_{ij}$} denote the mean and the variance of the $i$-th feature in the $j$-th class, respectively. Then the Fisher Score for the $i$-th 
feature is: 
\small\begin{equation}S_i = \frac{\sum\limits_{j} {n_j(\mu_{ij}-\mu_i)^2}}{\sum\limits_{j}{n_j\rho_{ij}^2}}\end{equation}\normalsize

\myemph{Pearson Correlation Coefficient}: It is used as a measure for quantifying linear dependence between a feature variable \mbox{$X_i$} and a target 
variable \mbox{$Y_k$}. It ranges from -1 (perfect anticorrelation) to 1 (perfect correlation) and is defined as:
\small\begin{equation}R_i = \frac{cov(X_i, Y_i)}{\sqrt{var(X_i)var(Y_i)}}\end{equation}\normalsize
where the covariance and variance of the variables are estimated from the training data.

\myemph{Mutual Information}: The Information Gain metric provides a method of measuring the dependence between the ith feature and the target classes 
\mbox{$\vec{c}=[c_1,c_2,\dotsc,c_k]$}, as the decrease in total entropy, namely \mbox{$IG(f_i, \vec{c})=H(f_i)- H(f_i|\vec{c})$}, where 
\mbox{$H(f_i)$} is the entropy of $f_i$ and \mbox{$H(f_i|\vec{c})$} the entropy of \mbox{$f_i$} after observing \mbox{$\vec{c}$}. High information 
gain indicates that the selected feature is relevant. IG has gained popularity due to its computational efficiency and simple interpretation. 
It has also been extended to account for feature correlation and redundancy. Other MI metrics are Gini Impurity and Minimum-Redundancy-Maximum-Relevance.

Filter models select features based on their statistical similarities to a target variable. Wrapper methods take a different approach and use the
preselected classifier as a way to evaluate the accuracy of the classification task for a specific feature subset. A wrapper algorithm consists of 
three components, namely a feature search component, a feature evaluation component and a classifier. At each step, the search component generates 
a subset of features which will be evaluated for the classification task. When the total number of features is small, it is possible to test all 
possible feature combinations. However, this approach, known as SUBSET, becomes quickly computationally intractable. 

Greedy search methods overcome this problem by using a heuristic rule to guide the subset generation. In particular, forward selection starts with an empty 
set and evaluates the classification accuracy of each feature separately. The best feature initializes the set. In the subsequent iterations, the current 
set is combined with each of the remaining features and the union is tested for its classification accuracy. The feature producing the best classification 
is added permanently to the selected features and the process is repeated until the number of features reaches a threshold or none of the remaining features 
improve the classification. On the other hand, backward elimination starts with all features. At each iteration, all features in the set are removed one by 
one and the resulting classification is evaluated. The feature affecting the classification the least, is removed from the list. Finally, bidirectional 
search starts with an empty set (expanding set) and a set with all features (shrinking set). At each iteration, first a feature is forward selected and 
added to the expanding set with the constraint that the added feature exists in the shrinking set. Then a feature is backward eliminated from the shrinking 
set with the constraint that it has not already been added in the expanding set. 

Many more strategies have been used to search the feature space, such as branch-and-bound, simulated annealing and genetic algorithms. 
Branch-and-bound uses depth-search to traverse the feature subset tree, pruning those branches which have worse classification score
than the score of an already traversed fully expanded branch. Simulated annealing and genetic algorithms encode the selected features in a 
binary vector. At each step, offspring vectors, representing different combinations of features, are generated and tested for their accuracy. 
A common technique for performance assessment is $k$-fold cross-validation. The training data are split into $k$ sets and the classification task
is performed $k$ times, using at each iteration one set as the validation set and the remaining $k$-1 sets for training. 

Filter methods are cheap, but selected features do not consider the biases of the classifiers. Wrapper methods select features 
tailored to a given classifier, but have to run the training phase many times, hence they are very expensive. Embedded methods combine the
advantages of both filters and wrappers by integrating feature selection in the training process. For example, pruning in Decision Trees and 
Rule-based Classifiers is a built-in mechanism to select features. In another family of classification methods, the change in the loss function
incurred by changes in the selected features, can be either exactly computed or approximated, without the need to retrain the model for each candidate 
variable. Combined with greedy search strategies, this approach allows for efficient feature selection (\myeg RFE/SVM, Gram-Schmidt/LLS). 
A third type of embedded methods are regularization methods and apply to classifiers where weight coefficients are assigned to features  (\myeg SVM 
or logistic regression). In this case, the feature selection task is cast as an optimization problem with two components, namely maximization 
of goodness-of-fit and minimization of the number of variables. The latter condition is achieved by forcing weights to be small or exactly zero. 
Features with coefficients close to zero are removed. 

Specifically, the feature weight vector is defined as:
\begin{equation}{\hat{\vec{w}}} = \argmin_{\vec{w}} c(\vec{w}, X) + penalty(\vec{w})\end{equation}
where c is the loss function, \mbox{$penalty(\vec{w})$} is the regularization term.
A well studied form of the regularization term is 
\begin{equation}penalty(\vec{w})=\lambda\parallel{\vec{w}}\parallel_{p}^{p}\end{equation} 
\myemph{Lasso regularization} uses \mbox{$p = 1$}, while
for \mbox{\myemph{Ridge} {$p = 2$}}. \myemph{Elastic net} combines the two:
\mbox{$penalty(\vec{w})=\lambda_1\parallel{\vec{w}}\parallel_{1} + \lambda_2\parallel{\vec{w}}\parallel_{2}^{2}$}

Many more feature selection algorithms and variations can be found in literature. Due to its significance in the classification task, feature selection 
and feature engineering in general, is a highly active field of research. For an in-depth presentation, the interested reader is referred to 
\cite{REF41}, \cite{REF42} and \cite{REF43}. Comprehensive reviews can be found in \cite{REF44} and \cite{REF45}.

\section{Ontologies}
The enormous amount of information available in the continuously expanding Web by far exceeds human processing capabilities. This gave rise to the 
question whether it is possible to build tools which will automate information retrieval and knowledge extraction from the Web repository. 
The Semantic Web emerged as a proposed solution to this problem. In its essence, it is an extension to the Web, in which content is represented in 
such a way that machines are able to process it and infer new knowledge from it. Its purpose is to alleviate the limitations of current knowledge engineering
technology with respect to searching, extracting, maintaining, uncovering and viewing information, and support advanced knowledge-based systems. 
Within the Semantic Web framework, information is organized in conceptual spaces according to its meaning. Automated tools search for 
inconsistencies and ensure content integrity. Keyword-based search is replaced by knowledge extraction through query answering. 

In order to realise its vision, the Semantic Web does not rely on ``exotic'' intelligent technology, where agents are able to mimic humans in
understanding the predominant HTML content. Rather it approaches the problem from the Web page side. Specifically, it requires Web pages to 
contain informative (semantic) annotations about their content. These semantics (metadata) enable software to process information without the 
need to ``understand'' it. The eXtensible Markup Language (XML) was a first step towards this goal. Nowadays, the Resource Description Framework (RDF), 
RDF Scheme (RDFS) and the Web Ontology Language (OWL) are the main technologies which drive the implementation of the Semantic Web.

In general, ontologies are the basic building blocks for inference techniques on the Semantic Web. As stated in W3C's OWL Requirements Documents 
\cite{REF46}:``An ontology defines the terms used to describe and represent an area of knowledge.'' Ontological \emph{terms} are concepts and
properties which capture the knowledge of a domain area. Concepts are organised in a hierarchy which expresses the relationships among them by means of
superclasses, representing higher level concepts and subclasses, representing specific (constrained) concepts. 
Properties are of two types: those that describe attributes (features) of the concepts, and those that introduce binary relations between the concepts.

In order to succeed in the goal to express knowledge in a machine-processable way, an ontology has to exhibit certain characteristics, namely
abstractness, preciseness, explicitness, consensus and domain specificity. An ontology is abstract when it specifies knowledge in a conceptual way. 
Instead of making statements about specific occurrences of individuals, it tries to cover situations in a conceptual way. Ontologies are expressed in 
a knowledge representation language which is grounded on formal semantics, \myie it describes the knowledge rigorously and precisely. Such semantics 
do not refer to subjective intuitions, nor are they open to different interpretations. Furthermore, knowledge is stated explicitly. Notions which are 
not directly included in the ontology are not part of the conceptualization it captures. In addition, an ontology reflects a common understanding of 
domain concepts within a community. In this sense a prerequisite of an ontology is the existence of social consensus. Finally, it targets a 
specific domain of interest. The more refined the scope of the domain, the more effective an ontology can be at capturing the details rather then
covering a broad range of related topics.

The most popular language for engineering ontologies is OWL \cite{REF47}. OWL (and latest OWL2) defines constructs, namely classes, associated properties and utility properties, which can be used to create domain vocabularies along with constructs for expressiveness (\myeg cardinalities, unions, intersections), thus enabling the modelling of complex and rich axioms. There are many tools available which support the engineering of  OWL ontologies (\myeg Prot\'eg\'e, TopBraid Composer) and OWL-based reasoning (\myeg Pellet, \mbox{HermiT}). Ontology engineering is an active topic and a growing number of fully developed domain and generic/upper ontologies are already publicly available, such as the Dublin Core (DC) \cite{REF48}, the Friend Of A Friend (FOAF) \cite{REF49}, Gene Ontology (GO) \cite{REF50}, Schema.org \cite{REF51}, to name a few. An extensive list of ontologies and ontology engineering methodologies have been recently published in Kotis et al. 2020 \cite{REF52}.

The Semantic Web is vast and combines many areas of research and technological advances. A comprehensive introduction can be found in \cite{REF53} and \cite{REF54}. The interested reader can find a detailed presentation of Semantic Web technologies in \cite{REF55}.

\section{Ontology-based Feature Selection}
The main research domain where ontologies have been employed to select features, is text classification, namely the task of assigning 
predefined categories to free-text documents based on their content. The continuous increase of volumes of text documents in the Internet, 
makes text classification an important tool for searching information. Due to their enormous scale in terms of the number of classes, training examples, features and feature dependencies, text classification applications present considerable research challenges. 

Elhadad \myetal \cite{REF56} use the WordNet \cite{REF57} lexical taxonomy (as an ontology) to classify Web text documents based on their semantic similarities. 
In the first phase, a number of filters are applied to each document to extract an initial vector of terms, called bag of words (BoW), 
which represent the document space. 
In particular, a Natural Language Processing Parser (NLPP) parses the text and extracts words in the form of tagged components (part of speech), 
such as verbs, nouns, adjectives, etc. Words which contain symbolic characters, non-English words and words which can be found in pre-existing stopping 
words lists, are eliminated. Furthermore, in order to reduce redundancy, stemming algorithms are used to replace words with equivalent morphological forms, 
with their common root. For example the words ``\emph{Fighting}'' and ``\emph{Fights}'' are replaced with the stemmed word ``\emph{Fight}''.
In the second phase, all words in the initial BoW are examined for semantic similarities with categories
in WordNet. Specifically, if a path exists, in the WordNet taxonomy, from a word to a WordNet category via a common parent (hypernym), 
then the word is retained, otherwise it is discarded. Once the final set of terms has been selected, the feature vector for each document is generated 
by assigning a weight to each term. Authors use the term frequency-inverse document frequency (TFIDF) statistical measurement, since it computes the 
importance of a term $t$, both in an individual document and in the whole training set. TFIDF is defined as:
\small\begin{equation}TFIDF(t) = TF(t) \times IDF(t) \end{equation}\normalsize
where \small\begin{equation}TF(t) =  \frac{Number\ of\ occurances\ of\ term\ t} {Total\ number\ of\ terms\ in\ doc} \end{equation}\normalsize
and \small\begin{equation}IDF(t) = \frac{log(Total\ Number\ of\ docs)}{Number\ of\ docs\ with\ term\ t} \end{equation}\normalsize
Effectively, terms which appear frequently in a document, but rarely in the overall corpus, are assigned larger weights.
Authors compared against the Principal Component Analysis (PCA) method and report superior classification results. However, they recognize that a limitation in their approach is 
that important terms that are not included in WordNet will be excluded from the feature selection.

Vicient \myetal \cite{REF58}, employ the Web to support feature extraction from raw text documents, which describe an entity (symbolized with \emph{ae}), according to 
a given ontology of interest. In the first step, the OpenNLP \cite{REF59} parser analyzes the document and detects potential named entities (\emph{PNE}) related to 
the \emph{ae}, as noun phrases containing one or more words beginning with a capital letter. A modified Pointwise Mutual Information (PMI) measure is used 
to rank the \emph{PNE} and identify those which are most relevant to the \emph{ae} according to some threshold. In particular, for each 
\mbox{$pne_i\in PNE$} probabilities are approximated by Web hit counts provided by a Web search engine, 
\begin{equation} NE_{score}(pne_i,ae) = \frac{hits(pne_i \& ae)}{hits(pne_i)}\end{equation}
In the second step, a set of subsumer concepts (\emph{SC}) is extracted from the retained named entities (\emph{NE}). To do so, the text is 
scanned for instances of certain linguistic patterns which contain each \mbox{$ne_i\in NE$}. Each pattern is used in a Web query and the resulting 
Web snippets determine the subsumer concepts representing the \emph{$ne_i$}. 
Next, the extracted \emph{SC} are mapped to ontological classes (\emph{OC}) from the input ontology. Initially, for each \emph{$ne_i$} all its 
potential subsumer concepts are directly matched to lexically similar ontological classes. If no matches are found then
WordNet is used to expand the \emph{SC} and direct matching is repeated. Specifically, the parents (\emph{hypernyms}) in the WordNet hierarchy of each subsumer 
concept \emph{$sc_i$} are added to \emph{SC}. In order to determine, which parent concepts are mostly relevant to the named entity \emph{$ne_i$}, 
a search engine is queried for common appearances of the \emph{ae} and the \emph{$ne_i$}. The returned Web snippets are used to determine which 
parent synsets of \emph{$sc_i$} are mostly related to \emph{$ne_i$}. Synsets in Wordnet are groupings of words from the same lexical category which are 
synonymous and express the same concept. Finally, a Web-based version of the PMI measure, defined as
\small\begin{equation} SOC_{score}(soc_i,ne_i,ae) = \frac{hits(soc_i \& ne_i \& ae)}{hits(soc_i \& ae)}\end{equation}\normalsize
is used to rank each of the extracted ontological classes (\emph{$soc_i$}), related to a named entity. The \emph{$soc_i$} with the highest score which
exceeds a threshold is used as annotation. The authors tested their method in the Tourism domain. For the evaluation, they compared precision (ratio of 
correct feature to retrieved features) and recall (ratio of correct features to ideal features) against manually selected features from human experts. They 
report 70-75\% precision and more than 50\% accuracy and argue that such results considerably reduce the human effort required to annotate textual resources.

Wang \myetal \cite{REF60} reduce the dimensionality of the text classification problem by determining an optimal set of concepts to identify document context (semantics). First, the document terms are mapped to concepts derived from a domain-specific ontology. For each set of documents of the same class, the 
extracted concepts are organized in a concept hierarchy. A hill-climbing algorithm \cite{REF61} is used to search the hierarchy and derive an optimal set 
of concepts which represents the document class. They apply their method to classification of medical documents and use the Unified Medical Language System 
(UMLS)\cite{REF62} as the underlying domain-specific ontology. UMLS query API is used to map document terms to concepts and to derive the concept hierarchy. 
For the hill-climbing heuristic, a frequency measure is assigned to each leaf concept node. The weight of parent nodes is the sum of the children' weights. 
Based on the assigned weights, a distance measure between two documents is derived, and used to define the fitness function. Test
documents undergo the same treatment and are classified based on the extracted optimal representative concepts. For their experiments the authors use a KNN classifier and report improved accuracy, but admit that an obvious limitation of their method is that it is only applicable in domains that have a fully developed ontology hierarchy.

Khan \myetal \cite{REF63} obtain document vectors defined in a vector space model. This is accomplished in terms of the following steps. First, after identifying all the words in the documents, they remove the stop-words from the word data base, creating a BoW. Next, a stemming algorithm is applied to assign each word to its respective root word.  Phrase frequency is estimated using a part of speech tagger (POS). Next, they apply the maximal frequent sequence (MFS) \cite{REF64} to extract the most frequent terms. MFS is a sequence of words that is frequent in a document collection and moreover it is not contained in any other frequent sequence \cite{REF64}. The final set of features is selected by examining similarities with ontology-based categories in WordNet [57] and applying a wrapper approach.  Using the TFIDF statistical measure weights are assigned to each term. Finally, the classifier is trained in terms of the Naive Bayes algorithm.

Abdollali \myetal 2019 \cite{REF65} also address feature selection in the context of classification of medical documents. In particular, they aim at
distinguishing clinical notes which reference Coronary Artery Disease (CAD) from those that do not. Similarly to \cite{REF60}, they use a query tool 
(MetaMap) to map meaningful expressions, in the training documents to concepts in UMLS. Since, their target is CAD documents, they only keep
concepts ``Disease or Syndrome'' and ``Sign or Symptom'' and discard the rest. The retained concepts are assigned a TFIDF weight to form the feature vector matrix which will be used in the classification. In the second stage, the Particle Swarm Optimization \cite{REF66} algorithm is used to select the optimal feature subset. The value of particles is initialized randomly by numbers in [-1, 1], where a positive number indicates an active feature while a negative value an inactive one. The fitness function for each particle is based on the classification accuracy, \myie:
\small{\begin{equation} Fittness(S) = \frac{TP+TN}{TP+TN+FP+FN}\end{equation}}\normalsize
where S represents the features set, TP and FP are the number of correctly and incorrectly identified documents and TN and FN the number of
correctly and incorrectly rejected documents. 10-fold cross validation is used to compute a particle's fitness value as the average of the accuracies 
of ten classification runs. The authors evaluated their method using five classifiers (NP, LSVM, KNN, DT, LR) and reported both significant reduction
of the feature space and improved accuracy of the classification in most of their tests.

Lu \myetal 2013 \cite{REF67} attempt to predict the probability of hospital readmission within 30 days after a heart failure, by means of the medication
list prescribed to patients during their initial hospitalization. In the first stage, the authors combine two publicly accessible drug ontologies, 
namely RxNorm \cite{REF68} and NDF-RT \cite{REF69}, into a tree structure, that represents the hierarchical relationship between drugs. The RxNorm ontology serves as drug thesaurus, while NDF-RT as drug functionality knowledge base. The combined hierarchy consists of six class levels. 
The top three levels correspond to classes derived 
from the Legacy VA class list in NDF-RT and represent the therapeutic intention of drugs. The fourth level represents the general active ingredients of 
drugs. The fifth level refers to the dosage of drugs and uses a unique identifier to match drugs to the most representative class in RxNorm (RXCUI). 
The lowest level, refers to the dose form of drugs and uses the local drug code used by different hospitals. Each clinical drug corresponds to a 
single VA class, a single group of ingredients and a single RxNorm class. In the second stage, a top-down depth-first traversal of the tree hierarchy is 
used to select a subset of nodes as features. For each branch, the nodes are sorted according to the information gain ratio (IGR(F) = IG(F)/H(F)).
The features in the ordered list are marked for selection one by one, while parent and child features with lower score are removed from the list. 
In order to evaluate their method, the authors use the Naive Bayes classifier and employ the area under the receiver operating characteristic
curve to evaluate its performance. Their experiments showed that the ontology guided feature selection outperformed the other 
non-ontology-based methods.

An important application of ontology-based feature selection algorithms is the selection of manufacturing processes. Mabkhot\myetal \cite{REF70} describe 
an ontology-based decision support system (DSS), which aims at assisting the selection of a suitable manufacturing process (MPS) for a new product. 
In essence, selected aspects of MPS are mapped to ontological concepts, which serve as features in rules used for case-based reasoning.
Traditionally, MPS has relied on expert human knowledge to achieve the optimal matching between design specifications, material characteristics and
process capabilities. However, due to the continuous evolution in material and manufacturing technologies and the increasing product complexity, this
task becomes more and more challenging for humans. The proposed DSS consists of two components, namely the ontology and the case-based reasoning 
subsystem (CBR). The purpose of the ontology is to encode all the knowledge related to manufacturing in a way which enables the reasoner to make a 
recommendation for a new product design. It consists of three main concepts, the manufacturing process (MfgProcess), the material (EngMaterial) and 
the product (EngProduct). The MfgProcess concept captures the knowledge about manufacturing in subconcepts, such as casting, molding, forming, 
machining, joining and rapid manufacturing. The properties of each manufacturing process are expressed in terms of shape generation capabilities, 
which describe the product shape features a process can produce, and range capabilities, which express the product attributes that can be met by 
the process such as dimensions, weight, quantity and material. The EngMaterial concept captures knowledge about materials, in terms of material type 
(\myeg metal, ceramic, etc) and material process capability (\myeg sand casting materials). The EngProduct concept encodes knowledge about 
products, defined in the form of shape features and attributes. The ontology facilitates the construction of rules, which relate manufacturing 
processes with engineering products, through the matching of product features and attributes, with process characteristics and capabilities. The semantic 
Web rule language (SWRL \cite{REF71}) has been used as an effective method to represent causal relations. The purpose of the CBR subsystem is to find the optimal 
product-to-process matching. It does so in two steps. First, it scans the ontology for a similar product. To quantify product similarity, appropriate 
feature and attribute similarity measures have been developed and human experts have been employed to assign proper weights to features and attributes. 
If a matching product is found then the corresponding process is presented to the decision maker, otherwise SWRL rule-based reasoning is used to find a 
suitable manufacturing process. Finally, the ontology is updated with the newly extracted knowledge. The authors presented a use case to demonstrate the 
usability and effectiveness of the proposed DSS and argue that in the future such systems will become more and more relevant.

In \cite{REF72}, Kang \myetal develop an ontology-based representation model to select appropriate machining processes as well as the corresponding inference rules. The ontology is quantified in terms of features, process capability with relevant properties, machining process, and relationships between concepts. A reasoning inference mechanism is applied to obtain the final set of processes for individual features. The final process with the highest contribution is determined by a procedure that matches the accuracy requirements of a specific feature with the capability of the candidate processes. The preceding machining process is then selected so that the precedence relationship constraint between the processes is met until no further precedent processes are required. The whole process selection scheme is neutral (i.e., general enough) in the sense that it does not depend on a specific restriction, and thus it constitutes a reusable platform.

Hat \myetal \cite{REF73} also apply ontology within the mechanical engineering domain, in particular the field of Noise, Vibration and Harshness (NVH). 
Similar to the previous work, authors map important aspects of noise identification to ontological concepts, which serve as features for reasoning.
They propose an ontology-based system for identifying noise sources in agricultural machines. At the same time, their method provides 
an extensible framework for sharing knowledge for noise diagnosis. Essentially, they seek to encode prior knowledge relating noise sound signals 
(targets) with vibrational sound signals (sources) in an ontology, equipped with rules, and perform reasoning to identify noise sources 
based on the characteristics of test input and output sound signals (parotic noise). In order to build the ontology, first, professional experience, 
literature and standard specifications were surveyed to extract the concepts related to NVH. The Prot\'eg\'e tool was used to convert the concept 
knowledge into an OWL ontology and implement the SWRL rules, which match sound source and parotic noise signals. The Pellet tool is employed for reasoning.
To quantify the signal correlations, the time signals are converted to the frequency domain and the values for seven common signal characteristics are 
calculated. Specifically, relation of the frequency of the parotic signal to the ignition frequency, peak frequency, Pearson Coefficient, frequency doubling, 
loudness, sharpness and roughness. The effectiveness of the method was demonstrated in a use case, where the prototype system correctly identified the main 
noise source. After improving the designated area the noise was significantly reduced. The authors argue that the continuous improvement in the 
knowledge base and rule set of the ontology model, has the potential to allow the design system to perform reasoning that simulates the thinking
process of the expert in the field of NVH.

Belgiu \myetal 2014 \cite{REF74} develop an ontological framework to classify buildings based on data acquired with Airborne Laser Scanning (ALS).
They followed five steps. Initially, they preprocessed the ALS point cloud and applied the eCognition software to convert it to raster data, which were
used to delineate buildings and remove irrelevant objects. Additionally, they obtained values for 13 building features grouped in four 
categories: extent features, which define the size of the building objects, shape features, which describe the complexity of building 
boundaries, height and slope of the buildings' roof. In the next step, human expert knowledge and knowledge available in literature was employed
to define three general purpose building ontology classes, independent of the application and the data at hand, namely
\begin{enumerate*}[label=\Alph*.]
\item \emph{Residential/Small Buildings}, \item \emph{Apartment/Block Buildings} and \item \emph{Industrial and Factory Buildings}. 
\end{enumerate*}
In order to identify the metrics which were mostly relevant to the identification of building types, a set with 45 samples was used to train a 
Random Forest classifier with 500 trees and $ \sqrt{m} $ features (m number of input features). The feature selection process identified 
slope, height, area and asymmetry as the most important features. The first three were modelled in the ontology with empirically 
determined thresholds by the RF classifier. Finally, building type classification was carried out based on the formed ontology. 
The classification accuracy was assessed by means of precision, recall and $F$-measure  
and the authors reported convincing results for class A while classes B and C had less accurate results. However, they argue that their method can prove 
useful for classifying building types in urban areas.

\par Finally, two interesting applications of ontology-based feature selection algorithms concern the recommendation systems (RS) and the information security/privacy research areas. In \cite{REF75}, Di Noia \myetal develop a filter-based feature selection algorithm by incorporating ontology-driven data summarization for linked-data (LD) based recommender system (RS).  The feature selection mechanism determines the k most important features in evaluating the similarity between instances of a given class on top of data summaries built with the help of an ontology. Two types of descriptors are employed namely, pattern frequency and cardinality descriptors. 
A pattern is defined as a schema using an RDF triple denoted as (C, P, D), where C and D are classes or datatypes, and P is a property that expresses their relationship. C is called the source type and D the target type. The patterns are used to generate data summarization from a knowledge graph-based framework. Each pattern is associated with a frequency that corresponds to the number of relational assertions from which the pattern has been extracted. Therefore, a pattern frequency descriptor can be viewed as a set of statistical measures. 
A cardinality descriptor encodes information about the semantics of properties as used within specific patterns and can be used in computing the similarities between these patterns. To obtain the cardinality descriptors, the authors extended the above-mentioned knowledge graph framework.
The LD and one or more ontologies are the inputs to the knowledge graph framework, while its outputs are: a type graph, a set of patterns along with the respective frequencies, and the cardinality descriptors.
To this end, the filtering-based feature selection consists of two main steps. First, the cardinality descriptors are implemented to filter out features (i.e., pattern properties) that correspond to properties connecting one target type with many source types. Second, the pattern frequency descriptors are applied to rank in a frequency-based descending order all features and select the top k features.

In \cite{REF76}, Guan \myetal studied the problem of mapping security requirements (SR) to security patterns (SP). Viewing the SPs as features, feature selection is set up to perform the above mapping procedure.  This selection is based on developing an ontology-based framework and a classification scheme. To accomplish this task, they described the SRs using four attributes namely, asset (A), threat (T), security attribute (SA), and priority (P).  The SRs are represented as rows in a two-dimensional matrix, where the columns correspond to the above attributes.  Then, the meaning of each SR is, for a given asset A, one or more threats Ts may threaten A by violating one or more attribute values of SA. In addition, each SR is to be fulfilled in a sequence according to the value of P during software development. Then, they generate complete and consistent SRs by eliciting values for the above attributes using the risk-based analysis proposed in \cite{REF77}. 
On the other hand, security patterns (SP) are described in terms of three attributes namely, Context that defines the conditions and situation in which the pattern is applicable, Problem that defines the vulnerable aspect of an asset, and Solution that defines the scheme that solves the security. 
To intertwin the above information they developed a two-level ontological framework using an OWL-based security ontology. The first level concerns the ontology-based description of SRs and the second the ontology-based description of SPs. These descriptions were carried out by quantifying mainly the risk relevant and annotating security related information. 
To this end, a classification scheme selects an appropriate set of SPs for each SR. The classification scheme is developed by considering multiple aspects such as lifecycle stage that organizes, architectural layer that organizes information from low to high abstraction level, application concept that partitions the security patterns according to which part of the system they are trying to protect, and threat type that uses the security problems solved by the patterns.

\section{Open Issues and Challenges}
Features show dependencies among each other and therefore, they can be structured as trees or graphs. Ontology-based feature selection in the era of knowledge graphs such as DBpedia, Freebase and YAGO, can be influenced by two issues \cite{REF78}:
\begin{enumerate}[label=\alph*.]
\item The large expansion of knowledge recorded in Wikipedia, from which DBpedia and YAGO has been created as reference sources for general domain knowledge needed to assist information disambiguation and extraction,
\item advancements in statistical NLP techniques, and the appearance of new techniques that combine statistical and linguistic ones. 
\end{enumerate}
An important and open issue in this domain is the linking of one document-mentioned entity to a particular KG's entity and they way it affects how other surrounding document entities are linked. Furthermore, it is more and more common nowadays to see an increasing number of inter-task dependencies being modeled, where pairs of tasks such as 
\begin {enumerate*}[label=\alph*)]
\item Named-Entity Recognition (NER) and Entity Extraction and Linking (EEL), 
\item WordSense Disambiguation (WSD) and EEL, or 
\item EEL and Relation Extraction and Linking(REL), are seen as interdependent. 
\end{enumerate*}
The combinatorial approach of those tasks will continue to exist and advance since it has been proven highly effective to the precision of the overall information/knowledge extraction process. Regarding the contributed communities in this area of research, related works have been conducted by the Semantic Web community as well as from others such as the NLP, AI and DB communities.  Works conducted by the NLP community focus more on unstructured input, while Database and Data Mining related works target more to semi-structured input \cite{REF78}. 

As abovementioned, ontologies play a key-role in feature selection. However, the engineering of ontologies, although has been  advancing in a fast pace over the last decade, it has not yet reached the status were consensus in domain-specific communities will deliver gold-standard ontologies for each case and application area. On the other hand, a number of issues and challenges related to the collaborative engineering of reused and live ontologies have been recently reported \cite{REF52}, indicating that this topic is still active and emerging. For instance, as far as concerns feature selection, different ontologies of the same domain used in the same knowledge extraction tasks will most probably result in different set of features selected (schema-bias). Furthermore, human bias in conceptualizing context during the process of engineering ontologies (in a top-down collaborative ontology engineering approach) will inevitably influence the feature selection tasks. Specifically, in the cases where large KGs (e.g., DBpedia) are used for knowledge extraction, such a bias is present in both conceptual/schema (ontology) and data (entities) levels. Debiasing KGs is a key challenge in the Semantic Web and KG community itself \cite{REF79}, and consequently in the domain of KG-based feature selection.\
\par Important challenges arise when ontology-based feature selection is applied to linked data (LD). LD appear to be one of the main structural elements of big data. For example, data created in social media platforms are mainly LD. LD appear to have significant correlations regarding various types of links and therefore, they possess more complex structure than the traditional attribute-valued data. However, they provide extra, yet valuable, information \cite{REF80}. The challenges of using ontology-based feature selection in LD concern the development of ontology-based frameworks to exploit complex relation between data samples and feature and how to use them in performing effective feature selection, and to evaluate the relevance of features without the guide of label information.
\par Another interesting research area is the real-time feature selection. The main difficulty in dealing with real-time feature selection is that both data samples and new features must be taken into account simultaneously. Most of the methods that exist in the literature rely on feature pre-selection or on feature selection without online classification \cite{REF81, REF82}. On the other hand ontology encoded in trees or knowledge graphs may provide some benefits such as solid representations of the current relations between features, which can be used to predict any possible relation between the current available features and the ones that are expected to arise in real-time processing tasks. Therefore, achieving real-time analysis and prediction for high-dimensional datasets remains a challenge. 
\par Finally, an important open issue to consider is scalability. Scalability quantifies the impact imposed by increasing the training data size on the computational performance of an algorithm in terms of accuracy and memory \cite{REF80, REF82}. The basics of feature selection and classification were developed before the era of big data. Therefore, most feature selection algorithms do not scale well on extremely high-dimensional data; their efficiency deteriorates quickly or is even computationally infeasible. On the other hand, scaling-up favors the accuracy of the model. Therefore, there is a trade-off between finding an appropriate set of features and the model’s accuracy. In this direction, the challenge is to define appropriate ontology-based relations between features in order to group them in such a way that the resulting set of features will be able to maintain acceptable model’s accuracy. 

\section{Conclusion}
This study provides an overview of ontology-based classification with emphasis on the feature selection process. The presented methodologies show
that ontologies can effectively uncover dominant features in diverse knowledge domains and can be integrated into existing feature selection
and classification algorithms. Specifically, in the context of text classification, domain-specific ontologies combined with the WordNet taxonomy, 
can be utilized to map terms in documents to concepts in the ontology, thus replacing specific term-based document features with abstract and generic
concept-based features. The latter capture the content of the text and can be used to train accurate and efficient classifiers. In the field
of mechanical engineering, ontologies can be employed to map human knowledge to concepts, that serve as features for case-based reasoning,
and support decision making, such as selection of manufacturing process or noise source identification. To this end, in the area of civil engineering, building type recognition can be facilitated by ontology. Although, this survey is by no means exhaustive, it demonstrates the broad applicability and feasibility of ontology-based feature extraction and selection.
Finally, certain open issues and challenges are discussed and a number of relevant problems are identified.

\bibliographystyle{plain}

\end{document}